\documentclass{article}

\usepackage{PRIMEarxiv}

\usepackage[utf8]{inputenc} 
\usepackage[T1]{fontenc}    
\usepackage{hyperref}       
\usepackage{url}            
\usepackage{booktabs}       
\usepackage{amsfonts}       
\usepackage{nicefrac}       
\usepackage{microtype}      
\usepackage{fancyhdr}       
\usepackage{graphicx}       
\graphicspath{{media/}}     

\usepackage{amssymb}
\usepackage{amsmath}
\usepackage{pdflscape} 

\usepackage{algorithm}
\usepackage{algpseudocode}
\usepackage{amsfonts}
\usepackage{tabularx} 
\usepackage{xcolor} 
\usepackage{booktabs} 
\usepackage{makecell}

\usepackage{hyperref}
\newcommand{\Figref}[1]{Figure~(\ref{#1})}
\newcommand{\Tabref}[1]{Table~(\ref{#1})}
\newcommand{\Secref}[1]{Section~(\ref{#1})}

\newcommand{\Algref}[1]{Algorithm~(\ref{#1})}

\usepackage{pifont}

\title{MOMA-AC: A Preference-Driven Actor-Critic
Framework for Continuous Multi-Objective
Multi-Agent Reinforcement Learning
}

\author{
  Adam Callaghan \\
  University of Galway\\
  Galway\\
  Ireland \\
  H91 TK33\\
   \And
  Karl Mason \\
  University of Galway\\
  Galway\\
  Ireland \\
  H91 TK33\\
   \And
  Patrick Mannion \\
  University of Galway\\
  Galway\\
  Ireland \\
  H91 TK33
}

\begin{document}
\maketitle

\begin{abstract}
This paper addresses a critical gap in Multi-Objective Multi-Agent Reinforcement Learning (MOMARL) by introducing the first dedicated inner-loop actor–critic framework for continuous state and action spaces: Multi-Objective Multi-Agent Actor–Critic (MOMA-AC). Building on single-objective, single-agent algorithms, we instantiate this framework with Twin Delayed Deep Deterministic Policy Gradient (TD3) and Deep Deterministic Policy Gradient (DDPG), yielding MOMA-TD3 and MOMA-DDPG. The framework combines a multi-headed actor network, a centralised critic, and an objective preference-conditioning architecture, enabling a single neural network to encode the Pareto front of optimal trade-off policies for all agents across conflicting objectives in a continuous MOMARL setting. We also outline a natural test suite for continuous MOMARL by combining a pre-existing multi-agent single-objective physics simulator with its multi-objective single-agent counterpart. Evaluating cooperative locomotion tasks in this suite, we show that our framework achieves statistically significant improvements in expected utility and hypervolume relative to outer-loop and independent training baselines, while demonstrating stable scalability as the number of agents increases. These results establish our framework as a foundational step towards robust, scalable multi-objective policy learning in continuous multi-agent domains.\end{abstract}


\section{Introduction}
Reinforcement Learning (RL) has emerged as a powerful framework for tackling sequential decision-making problems across a wide range of domains, including robotic control, autonomous vehicles, and large language models \cite{ELALLID20227366, wang2024reinforcement, singh2022reinforcement}. RL algorithms learn effective policies by interacting with an environment, exploring different actions and adapting based on environmental feedback. However, real-world scenarios are often complex. They may involve multiple decision-making entities, each with cooperative, competitive, or mixed incentives, and objectives that are inherently multi-faceted and conflicting. For example, in a collaborative manufacturing process, agents may be tasked with meeting production targets under time constraints while also minimising energy consumption. While such settings have led to the development of Multi-Agent Reinforcement Learning (MARL) and Multi-Objective Reinforcement Learning (MORL) as distinct subfields, both can be viewed as constrained instances of a more general problem. This motivates the unification of these perspectives into the more expressive framework of MOMARL.

In recent years, both MARL and MORL have matured as independent research areas, with notable advances in algorithmic design and theoretical analysis \cite{hayes2022practical, canese2021multi}. However, their intersection, MOMARL, remains comparatively under-explored. Several studies have attempted to address MOMARL challenges in practical domains such as traffic signal control, economic dispatch, and smart home energy management \cite{schester2021automated, lu2022multi}. Yet, these efforts often collapse the problem to suit existing MARL frameworks by reducing vector-valued rewards into scalar objectives a priori, yielding only a single policy per training run and failing to capture the diversity of trade-offs inherent in multi-objective problems \cite{radulescu2020recent}.

Likewise, direct extensions of MORL algorithms to multi-agent settings via decentralised learning, where each agent trains independently, overlook critical challenges unique to MARL. Without centralised coordination or shared training signals, agents who learn independently cannot adequately cope with the non-stationarity introduced by co-learners, nor can they address partial observability and the credit assignment problem. These limitations often result in degraded performance or emerging behaviours characterised by agents that contribute minimally or ineffectively to overall team objectives. While some early works have proposed dedicated MOMARL algorithms, they remain constrained to discrete state or action spaces, limiting their utility in real-world applications that require continuous control, fine-grained coordination, and scalable policy learning.

To bridge the methodological gap in MOMARL, we propose MOMA-AC, the first inner-loop actor–critic framework specifically designed for continuous state and action spaces that unifies multi-objective optimisation with multi-agent coordination. We instantiate the framework using two widely adopted single-objective algorithms—Twin Delayed Deep Deterministic Policy Gradient (TD3) \cite{fujimoto2018addressing} and Deep Deterministic Policy Gradient (DDPG) \cite{lillicrap2015continuous}—resulting in MOMA-TD3 and MOMA-DDPG. Both variants adopt centralised training with decentralised execution (CTDE) to support stable, scalable learning. A centralised critic is trained to estimate vector-valued $Q$-functions over joint state–action pairs, capturing the reward structure across multiple objectives. In parallel, decentralised actors use a multi-headed architecture, with each agent’s policy executed locally.

MOMA-AC follows an inner-loop design in which both actor and critic are conditioned on scalarisation preference weights, enabling a single actor network to represent a family of policies aligned with different scalarisations and thus approximate the Pareto front without separate training runs or multiple networks. This contrasts with approaches that produce one solution per scalarisation or fail to coordinate agent behaviour in a multi-objective setting.

To evaluate our approach, we construct a coherent testbed for continuous MOMARL by combining two previously separate frameworks: the multi-agent MuJoCo simulator (MaMuJoCo) \cite{gymnasium_robotics2023github} and the vector-valued reward structure from the mo-gymnasium suite \cite{felten_toolkit_2023}. While each independently addresses multi-agent or multi-objective extensions, their integration enables high-fidelity, continuous control environments with multi-objective feedback, making them a natural choice for benchmarking algorithms in the MOMARL setting. We focus on decomposed locomotion tasks in which agents control distinct segments of a shared robot and must coordinate to optimise conflicting objectives, such as minimising energy consumption while maximising displacement or velocity. This domain presents a realistic, high-dimensional testbed for evaluating the scalability and effectiveness of our proposed method.

Through comprehensive empirical evaluations, we demonstrate that MOMA-AC consistently outperforms two conventional baselines: a decentralised variant in which each agent independently trains a multi-objective TD3 policy, and an outer-loop baseline in which MA-TD3 \cite{lei2024multi} is applied iteratively across a sequence of fixed scalarisation weights. Across a broad spectrum of preferences, MOMA-TD3 achieves substantial gains in expected utility (EUM) and Pareto coverage metrics, highlighting its ability to learn diverse and efficient trade-off policies.

Furthermore, we show that the framework scales robustly with the number of agents, maintaining performance as more agents are introduced to the system.

Lastly, we investigate the backbone algorithm choice by comparing MOMA-TD3 and MOMA-DDPG. In particular, MOMA-TD3 retains TD3’s dual-critic architecture, delayed policy updates, clipped exploration noise and noisy target actions. \Secref{sec:utility_overestimation} shows how these backbone choices continue to play a role in mitigating overestimation bias during utility estimation.

Our key contributions are summarised as follows:
\begin{enumerate}
    \item We introduce MOMA-AC, the first inner-loop multi-objective multi-agent RL framework designed specifically for continuous state and action spaces, capable of approximating Pareto fronts via scalarisation-based training.
    \item We propose a novel architectural combination of multi-headed actors and centralised critics that supports CTDE training and addresses key coordination and utility estimation challenges in MOMARL.
    \item We conduct a thorough scalability analysis of instantiations of the framework, showing that MOMA-TD3 and MOMA-DDPG maintain robust performance as the number of agents increases in cooperative robotic control tasks.
    \item We empirically demonstrate that the dual-critic mechanism remains essential for mitigating utility overestimation in vector-valued reward settings.
\end{enumerate}

The remaining sections of the paper are structured as follows; \Secref{sec:background} reviews relevant literature in MORL, MARL and their intersection. \Secref{sec:framework} formalises the problem setup and underlying mathematical framework. \Secref{sec:methodology} presents the MOMA-AC framework in detail, including architectural choices and training methodology. \Secref{sec:experiments} describes the experimental setup. \Secref{sec:results} reports results and analysis. \Secref{sec:future_work} provides future research directions. Lastly, \Secref{sec:conclusion} provides a concluding summary of the work.

\section{Background}\label{sec:background}

This section introduces the concepts required to follow MOMA‑AC.  
We first recap single‑agent reinforcement learning, then outline its two orthogonal extensions—multi‑agent learning and multi‑objective learning—before describing their joint formulation as well as identifying the gap in the literature that our work addresses.

\subsection{Reinforcement Learning}\label{sec:rl}

Reinforcement learning casts sequential decision‑making as a Markov decision process \cite{sutton1998reinforcement}.  
At each step, an agent observes a state \(s\), selects an action \(a\), and receives a scalar reward \(r\).  
The objective is to learn a policy that maximises the expected discounted return over time.  
This objective is commonly expressed via value functions, most notably the state–action value \(Q(s,a)\), which estimates the expected return of taking action \(a\) in state \(s\) and following the optimal policy thereafter.  
Temporal-difference methods are often employed to learn these value functions by bootstrapping from predicted next-state values, updating estimates from partial returns rather than waiting for complete trajectories.

Early applications explored discrete state and action spaces and stored \(Q\)-values in tabular form, updating them iteratively using Q‑learning \cite{watkins1992q}.  
However, tabular methods do not scale to high‑dimensional or continuous domains due to the curse of dimensionality.  
Deep Q‑Networks (DQN) addressed this limitation by replacing tables with neural approximators and stabilising training through experience replay and target networks \cite{mnih2013playing}.

For continuous control, actor–critic architectures have become prevalent.  
These architectures simultaneously approximate the value function \(Q_\theta(s,a)\) and a policy \(\pi_\phi\).  
DDPG \cite{lillicrap2015continuous} introduced an off‑policy variant suitable for continuous actions, while TD3 \cite{fujimoto2018addressing} improved stability by adding a twin‑critic scheme and delayed policy updates.

With the rapid growth of the field, many competitive algorithms have been presented in recent years, such as Soft Actor–Critic (SAC) \cite{haarnoja2018soft}, which maximises a combination of reward and entropy to encourage exploration; Proximal Policy Optimisation (PPO) \cite{schulman2017proximal}, which constrains policy updates via a clipped surrogate objective to maintain stability; and evolutionary‑hybrid approaches such as Evolutionary RL (ERL) \cite{khadka2018evolution}, where gradient‑based learning is integrated with population‑based search.  

While these developments have made RL robust and scalable, they are intrinsically designed for single‑objective, single‑agent problems.  
Extending them to accommodate multi‑agent interactions or multiple objectives demands additional structural and algorithmic innovations, as discussed in the following sections.

\subsection{Multi‑Agent Reinforcement Learning}\label{sec:marl}

Although many control problems are framed in a single-agent context,
real-world environments are often inherently multi-agent. Agents act alongside others whose simultaneous actions influence the environment and, in turn, their own outcomes. This interdependence arises in domains such as autonomous driving \cite{schester2021automated,bhalla2020deep}, energy scheduling \cite{kofinas2018fuzzy}
and robot coordination \cite{peng2021facmac, gu2023safe}. MARL extends the reinforcement learning paradigm
to account for such interactions, enabling agents to learn effective behaviours
in environments influenced by other learning agents.

Such extensions raise three challenges \cite{busoniu2008comprehensive}.  
\emph{(i) Non‑stationarity:} every policy update by one agent changes the effective environment for the rest, violating the Markov and stationarity assumptions of standard RL.  
\emph{(ii) Partial observability:} agents typically perceive only local information rather than the full global state.  
\emph{(iii) Credit assignment:} when a team receives a joint reward, it is unclear how much each agent contributed.

A core design goal in MARL is to enable decentralised execution, where agents act independently, without requiring constant global coordination. For example, in autonomous driving it is impractical for each vehicle to require continuous access to the internal states and control actions of all other vehicles.
 Early methods such as Independent Q-Learning (IQL) attempted to avoid this by treating other agents as part of the environment. While computationally simple, this approach often led to instability due to the unpredictable behaviour of co-learners.
A more effective framework, CTDE, was later developed to address this limitation \cite{lowe2017multi}. 

Under CTDE, agents are trained using centralised information, such as the joint state and actions of all agents, but deploy policies based solely on local observations. Actor-critic architectures are naturally suited to this structure \cite{liang2025review}: a centralised critic estimates joint $Q$-values, while each agent’s actor is conditioned only on its own local input. Extensions of single-agent actor-critic algorithms, including DDPG and TD3, have been adapted to this setting resulting in methods like MADDPG and MATD3 that form the basis of many continuous-action MARL approaches \cite{lowe2017multi}.

To address the credit assignment problem, algorithms such as QMIX \cite{rashid2020monotonic} factorise the joint Q-function into individual components using a mixing network, while QTRAN \cite{son2019qtran} improves performance with fewer mathematical assumptions. Such decompositions allow agents to learn local policies while maintaining consistency with global reward signals.

Although MARL permits heterogeneous agent policies, standard formulations assume scalar-valued rewards. When agents must reason over explicit trade-offs (e.g., speed versus energy efficiency), this scalar assumption breaks down. In such cases MARL must be extended to support multi-objective reasoning, motivating its unification with MORL.

\subsection{Multi‑Objective Reinforcement Learning}\label{sec:morl}

Traditional reinforcement learning seeks to maximise a single scalar reward, an approach that suffices when task priorities reduce cleanly to one metric.  In many real‑world domains, the agent must reconcile several, often conflicting, objectives.  An autonomous vehicle, for instance, must trade off between speed, fuel efficiency, and passenger safety; a robotic arm may weigh positional accuracy against energy consumption.  Collapsing these criteria into a fixed linear scalarisation (utility), risks masking key trade‑offs and producing biased or brittle behaviour \cite{hayes2022practical,vamplew2022scalar}.

MORL addresses this limitation by representing the reward as a vector \(\mathbf{r}\in\mathbb{R}^d\), preserving feedback for each objective. The learning target then shifts from a single optimal policy to a set of Pareto-optimal policies. These are policies whose return vectors are mutually non-dominated, enabling downstream selection according to situational preferences.

Outer-loop methods treat each preference weight \( \ \boldsymbol{\omega} \), as a separate scalar task optimising them one at a time with standard RL machinery. Although conceptually simple, this approach can be sample-inefficient as it often requires complete retraining of a single-objective RL algorithm for each new preference. MORL/D \cite{felten2024multi} exploits neighbourhood relations between nearby weights to share information across tasks. PGMORL \cite{xu2020prediction} selectively trains only on the most promising weights and predicts solutions for unseen ones. GPI-LS \cite{alegre2023sample} prioritises which preferences and experiences to train on, using generalised policy improvement to incrementally build an effective coverage set.

Inner‑loop methods instead condition a single policy on \(\boldsymbol{\omega}\), amortising learning across weightings.  Successful applications of this can be seen through RL approaches such as CAPQL \cite{lu2023multi} and more recently through hybrid approaches such as MO‑ERL \cite{callaghan2025extending} and MOPDERL \cite{tran2023two}, which have been applied to multi-objective robotic control tasks.  While this method can achieve markedly higher sample efficiency it also requires careful design to ensure Pareto coverage and training stability.

Despite this progress, MORL research has largely focused on isolated agents.  Existing methods therefore overlook the non‑stationarity, partial observability, and credit assignment issues that arise when multiple learners interact.  Bridging this gap requires algorithms that combine preference‑conditioned optimisation with the coordination mechanisms of multi‑agent reinforcement learning, a synthesis addressed in the following section.

\subsection{Multi‑Objective Multi‑Agent Reinforcement Learning}\label{sec:momarl}

Many real-world systems—such as cooperative multi-agent path planning \cite{tao2024multi}, microgrid energy resource scheduling \cite{abid2024novel}, and residential appliance scheduling \cite{lu2022multi}—require both intra-agent trade-off handling (MORL) and inter-agent coordination (MARL). Yet work at this intersection remains sparse. In these settings, agents must pursue their own conflicting objectives while adapting to the behaviours and goals of others.

This intersection defines multi-objective multi-agent reinforcement learning (MOMARL): agents interact in a shared environment under vector-valued rewards and must reconcile individual preferences with joint dynamics. The challenge is compounded—coordination under partial observability and non-stationarity, together with optimisation across multiple, often conflicting objectives \cite{ruadulescu2024world}.

Early approaches to MOMARL often sidestepped its full complexity by scalarising vector rewards with fixed weights and applying existing MARL algorithms \cite{mannion2016multi}. While tractable, this approach produced only a single policy per preference and lacked mechanisms for trade-off exploration or generalisation. Furthermore, many methods relied on fully decentralised learning, leading to instability in the face of non-stationarity and limited coordination.

Recent works have begun to explore multi-policy learning in multi-agent settings. Tunable MODQN \cite{lu2022multi} addresses discrete-action domains by conditioning policies on sampled preference weights, while MO-MIX \cite{hu2023mo} incorporates mixing logic to tackle credit assignment in cooperative multi-agent tasks, though it is also limited to discrete state and action spaces. Algorithms such as MO-AIM \cite{dixit2023learning} and MOAVOA-MADDPG \cite{abid2024novel} extend learning to continuous-action MOMA frameworks for asynchronous agents; however, their outer-loop nature requires training separate policies for fixed preferences, lacking a unified, preference-conditioned solution capable of covering the Pareto front in a single training run.

\begin{table}[t]
\centering
\caption{Summary of related algorithms relative to MOMA-AC.}

\begin{tabularx}{\columnwidth}{lcccc}
\hline  
\textbf{Algorithm} & \makecell{\textbf{Multi-}\\\textbf{Agent}}  & \makecell{\textbf{Multi-}\\\textbf{Obj.}} & \textbf{Inner-Loop} & \makecell{\textbf{Cont. State}\\\textbf{\& Action}} \\
\hline
TD3 \cite{fujimoto2018addressing} & \ding{55} & \ding{55} & \ding{55} & \ding{51} \\
MADDPG/MATD3 \cite{lowe2017multi} & \ding{51} & \ding{55} & \ding{55} & \ding{51} \\
CAPQL \cite{lu2023multi} & \ding{55} & \ding{51} & \ding{51} & \ding{51} \\
MORL/D \cite{felten2024multi}& \ding{55} & \ding{51} & \ding{55} (outer) & \ding{51} \\
MO-MIX \cite{hu2023mo} & \ding{51} & \ding{51} & \ding{51} & \ding{55} (disc. action) \\
MO-AIM \cite{dixit2023learning}& \ding{51} & \ding{51} & \ding{55} (outer) & \ding{51} \\
Tunable MODQN \cite{lu2022multi} & \ding{51} & \ding{51} & \ding{51} & \ding{55} (disc. action) \\
MOAVOA-MADDPG \cite{abid2024novel} & \ding{51} & \ding{51} & \ding{55} (outer) & \ding{51} \\
MOMA-AC (ours) & \ding{51} & \ding{51} & \ding{51} & \ding{51} \\
\hline
\end{tabularx}
\footnotetext[1]{MO-MIX supports continuous state spaces but only discrete actions in its current implementation.}
\label{tab:algorithm_comparison}
\end{table}

To date, no existing algorithm satisfies three essential criteria for effective MOMARL in continuous domains: (i) support for continuous state and action spaces; (ii) decentralised execution of preference-conditioned policies; and (iii) centralised coordination via actor-critic training. Our proposed method, MOMA-AC, addresses this gap by extending the actor-critic framework with a multi-headed actor architecture, centralised critics, and utility-conditioned training. This enables scalable and stable learning of diverse, Pareto-efficient policies across agents in high-dimensional, continuous environments.

\section{Mathematical Framework}
\label{sec:framework}
We now formalise the mathematics underpinning MOMA-AC. We begin with a multi-objective Markov decision process (MOMDP) on continuous state and action spaces with vector-valued rewards, and then extend to the cooperative multi-agent case.

\subsection{MOMDP preliminaries}
A MOMDP is a tuple
\[
\mathcal{M} = (\mathcal{S},\mathcal{A},\mathcal{P},\mathbf{r},\gamma,\rho_0,\Omega),
\]
where $\mathcal{S}$ and $\mathcal{A}$ are continuous state and action spaces; $\mathcal{P}:\mathcal{S}\times\mathcal{A}\to\Delta(\mathcal{S})$ is the transition kernel; $\mathbf{r}:\mathcal{S}\times\mathcal{A}\to\mathbb{R}^d$ is a \emph{vector}-valued reward function; $\gamma\in[0,1)$ is the discount; $\rho_0\in\Delta(\mathcal{S})$ is the initial-state distribution; and $\Omega\subseteq\Delta^{d-1}=\{\boldsymbol{\omega}\in\mathbb{R}^d_{\ge 0}:\sum_{j=1}^d \omega_j=1\}$ is the set of preference weights over objectives.

\subsection{Multi-Objective Learning}

\paragraph{Outer-loop}
For each preference vector $\boldsymbol{\omega}\in\Omega$, define the scalarised reward
\[
r_{\boldsymbol{\omega}}(s,a) \;=\; \boldsymbol{\omega}^\top \mathbf{r}(s,a).
\]
For fixed $\boldsymbol{\omega}$, the scalar return of a policy $\pi$ from $s_0\sim\rho_0$ is
\[
J^\pi_{\boldsymbol{\omega}} \;=\; \mathbb{E}_{\,s_0 \sim \rho_0,\;\pi}\!\left[\sum_{t=0}^{\infty}\gamma^t\, r_{\boldsymbol{\omega}}(s_t,a_t)\right].
\]

Outer-loop methods seek to learn the family of policies $\{\pi^*_{\boldsymbol{\omega}}\}_{\boldsymbol{\omega}\in\Omega}$, each optimal for its scalarised objective $J^\pi_{\boldsymbol{\omega}}$.

\paragraph{Inner-loop}
Inner-loop methods differ in that they learn a \emph{single} policy $\pi$, whose action selection is now inherently conditioned by the preference over objectives described by $\boldsymbol{\omega}$. Thus we have $a = \pi(s, \boldsymbol{\omega})$. For such a policy, the induced set of vector returns is
\[
F_\pi=\Big\{\, \mathbb{E}_{\,s_0\sim\rho_0,\;\pi(\cdot, \boldsymbol{\omega})}\!\Big[\sum_{t=0}^\infty \gamma^t\,\mathbf r(s_t,a_t)\Big] \;|\ \boldsymbol{\omega}\in\Omega \Big\}.
\]
We define the coverage set as the non-dominated subset
\[
\mathrm{Cov}(\pi)\;=\;\Big\{\;\mathbf{u}\in F_\pi\;\Big|\;\nexists\,\mathbf{v}\in F_\pi\ \text{s.t.}\ \mathbf{v}\succeq \mathbf{u}\ \text{and}\ \mathbf{v}\neq \mathbf{u}\;\Big\},
\]
where $\mathbf{v}\succeq\mathbf{u}$ denotes componentwise inequality $v_j\ge u_j$ for all $j$. 

While outer-loop methods may select the best policy for each preference weight by comparing the policies' scalarised returns under said weight, inner-loop policies cannot be cherry-picked at the preference level. Therefore we must compare two conditional inner-loop policies based on their coverage-set quality. Often this is done by applying scalarisation metrics to the coverage set such as hypervolume and expected utility.

\subsection{Multi-Objective Multi-Agent Dec-POMDP}
We model MOMARL as a multi-objective decentralised partially observable Markov decision process (MO-Dec-POMDP) with multiple agents, local observations, and vector-valued rewards:
\[
\mathcal{G}
=
\Big(
\mathcal{S},\{\mathcal{A}_{i}\}_{i=1}^n, \mathcal{P},\mathbf{r},\{\mathcal{O}_{i}\}_{i=1}^n,\gamma,\rho_0,\Omega
\Big),
\]
where $\mathcal{A}_{i}$ is agent $i$’s continuous action space, $\mathcal{O}_{i}$ its observation space, and the remaining symbols are as above. The joint action and observation spaces are
\[
\mathcal{A}=\prod_{i=1}^n \mathcal{A}_{i},\qquad
\mathcal{O}=\prod_{i=1}^n \mathcal{O}_{i}.
\]
At each time step $t$, for a given weight $\boldsymbol{\omega}\in\Omega$, the environment is in state $s_t\in\mathcal{S}$. Each agent $i$ receives a local observation $o_{i_t}\in\mathcal{O}_{i}$ and selects an action
\[
a_{i_t}=\pi_{i}(o_{i_t},\boldsymbol{\omega})\in\mathcal{A}_{i}.
\]

Let $o_t=(o_{1_t},\ldots,o_{n_t})\in\mathcal{O}$ and $a_t=(a_{1_t},\ldots,a_{n_t})\in\mathcal{A}$. The joint action is applied, the environment transitions to $s_{t+1}\sim\mathcal{P}(\cdot\,|\,s_t,a_t)$, and the team receives $\mathbf{r}(s_t,a_t)\in\mathbb{R}^d$.
Thus, the $\boldsymbol{\omega}$–conditioned joint policy is
\[
\pi(o,\boldsymbol{\omega})=\big(\pi_{1}(o_{1},\boldsymbol{\omega}),\ldots,\pi_{N}(o_{n},\boldsymbol{\omega})\big).
\]

\subsection{Value functions and learning objective}
Given a conditioned policy $\pi(\cdot,\boldsymbol{\omega)}$, the value of the policy in state $s$ is
\[
\textbf{V}^{\pi(\cdot,\boldsymbol{\omega)}}(s)=
\mathbb{E}
\!\left[
\sum_{t=0}^{\infty}\gamma^t\, \mathbf{r}(s_t,a_t)
\ \middle|\,\ s_0=s, a_t{=}\pi(s_t,\boldsymbol{\omega})
\right]
\in\mathbb{R}^d,
\]

and the value of the policy is
\[
\textbf{V}^{\pi(\cdot, \boldsymbol{\omega})} = \textbf{V}^{\pi(\cdot, \boldsymbol{\omega})}(s_0), \qquad s_0\sim\rho_0.
\]
A preference vector allows for a natural scalarisation of the vector valued $\textbf{V}^{\pi(\cdot, \boldsymbol{\omega})}$ through the use of a utility function $U:\mathbb{R}^d \times\Omega\rightarrow\mathbb{R}$.

We define:
\[
U(\textbf{V}^{\pi(\cdot, \boldsymbol{\omega})}) =\boldsymbol{\omega}^\top \textbf{V}^{\pi(\cdot, \boldsymbol{\omega})}.
\]

Given a joint policy, the vector-valued state–action value is
\[
\textbf{Q}^{\pi}(s,a,\boldsymbol{\omega})
=
\mathbb{E}_{\pi(\cdot,\boldsymbol{\omega})}
\!\left[
\sum_{t=0}^{\infty}\gamma^t\, \mathbf{r}(s_t,a_t)
\ \middle|\ s_0{=}s,a_0{=}a
\right]
\in\mathbb{R}^d.
\]
 The corresponding \emph{scalarised} value is
\[
Q^{\pi}_{\boldsymbol{\omega}}(s,a)
=
\boldsymbol{\omega}^\top \textbf{Q}^{\pi}(s,a,\boldsymbol{\omega}).
\]

During training, the critic approximates $\textbf{Q}^{\pi}(s,a,\boldsymbol{\omega})$, and the actor is updated to maximise
$\mathbb{E}\!\left[\,Q^{\pi}_{\boldsymbol{\omega}}\!\big(s,\pi(s,\boldsymbol{\omega})\big)\right]$
over sampled $\boldsymbol{\omega}\sim\Omega$. 

Note:
\[
\mathbb{E}_{\,s_0 \sim \rho_0}\!\left[\,Q^{\pi}_{\boldsymbol{\omega}}\!\big(s_0,\pi(s_0,\boldsymbol{\omega})\big)\right] = U(\textbf{V}^{\pi(\cdot, \boldsymbol{\omega})}),
\]
thus the objective of the actor can be seen as the optimisation of the utility for all preference weights $\boldsymbol{\omega}$.

\section{Methodology}
\label{sec:methodology}
In this section we introduce MOMA-AC, the first multi-agent actor-critic framework designed for continuous control in multi-objective environments. MOMA-AC integrates CTDE, preference-conditioned policies, vector-valued rewards, and utility-based optimisation to learn a diverse set of solutions aligned with user-defined objectives for a team of co-operative agents. We instantiate this framework via two well-known actor critic algorithms; MOMA-TD3 and MOMA-DDPG.

MOMA-AC builds on three core components:
\begin{itemize}
    \item A centralised critic estimating vector-valued Q-functions using joint observations and actions.
    \item A single multi-headed actor encoding decentralised policies for all agents.
    \item A sampling and conditioning strategy for preference vectors to support coverage over the objective space.
\end{itemize}

\subsection{Centralised Critic}
The centralised critic serves as the primary mechanism for agent coordination and information sharing. The critic receives the  joint observation \(o=(o_1,...,o_n)\), joint action \(a=(a_1,...,a_n)\), and preference weight \(\boldsymbol{\omega}\) and in turn estimates the vector‑valued return \(\textbf{Q}(o,a,\boldsymbol{\omega})\).  
Because we operate in a fully cooperative setting, the environment provides a single, shared reward vector, so the critic estimates the team return conditioned on the joint observation, joint action, and preference weights.

Mirroring the backbone algorithm \cite{fujimoto2018addressing}, MOMA-TD3 uses two independently parametrised critics \(Q_{\theta_1}\) and \(Q_{\theta_2}\). At each update, the bootstrap target takes the pointwise minimum of the two target critics, see Equations \eqref{bootstrap target} and \eqref{eq:critic_target}.  In contrast, MOMA-DDPG maintains a single critic and forms the target with that critic alone.

To update the critic we first sample transitions $(o, a, \boldsymbol{\omega}, \mathbf{r}, o')$ from the replay buffer, where $o'$ is the joint observation after action $a$ was performed in $o$.  The joint next action is then produced using the multi-headed actor target parametrised by $\bar\phi$:
\begin{equation}
\label{eq:actor_target}
    a' = \bigl(a'_{1},\ldots,a'_{n}\bigr), \qquad
a'_{i} = \pi_{i,\bar\phi}\!\bigl(o'_{i}, \boldsymbol{\omega}\bigr) + \epsilon_{i},
\end{equation}

\begin{figure}[h]
    \centering
    \includegraphics[trim={0.5cm 11.5cm 0cm 1.2cm}, clip]{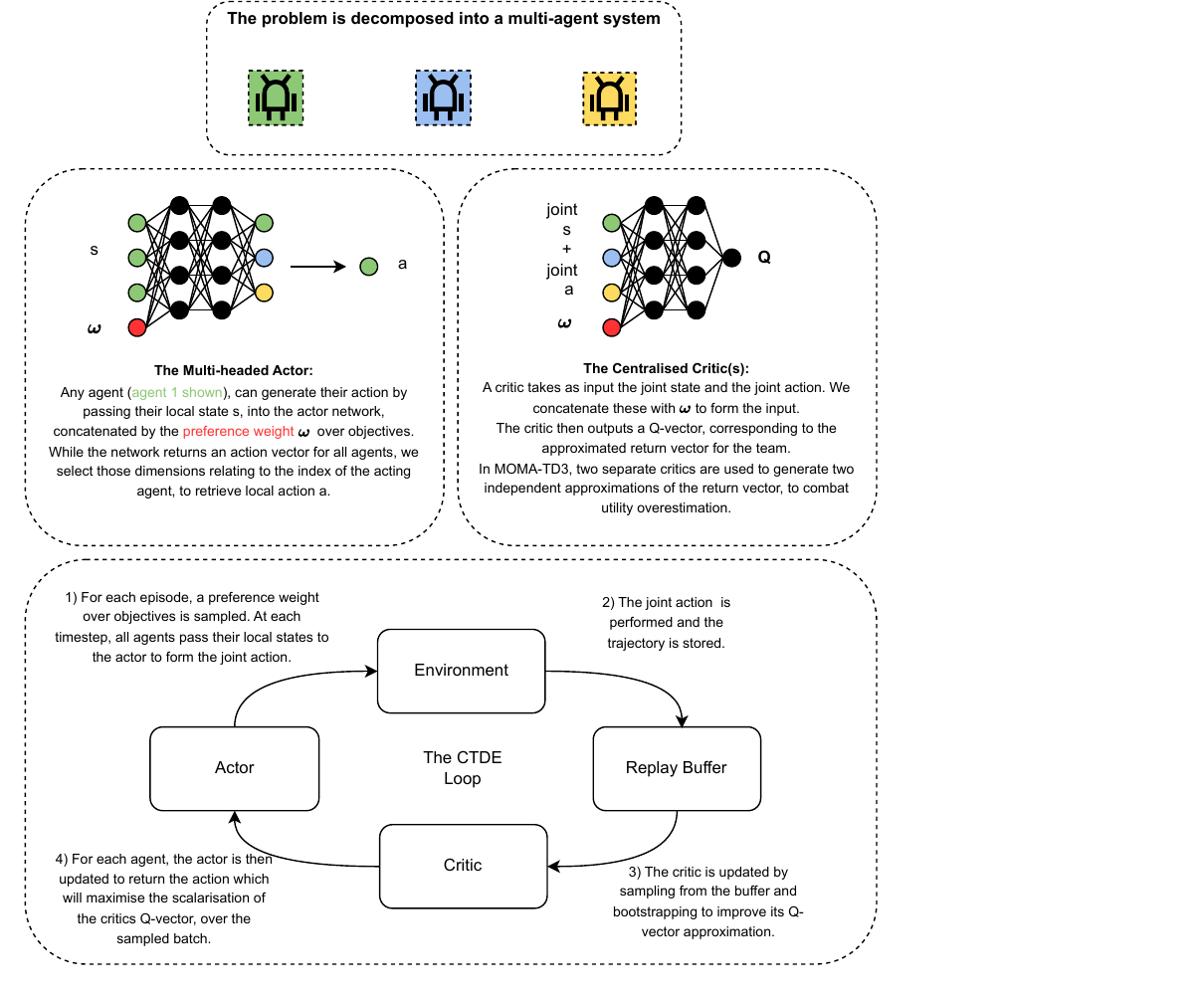}
    \caption{Architecture and training flow of the MOMA-AC framework for a three agent decomposition task.
A sampled preference weight $\boldsymbol{\omega}$ conditions both the multi-headed actor and the centralised critic.
Local observations and $\boldsymbol{\omega}$ form decentralised actions that interact with the environment, while the critic learns vector-valued Q-functions over joint states and actions.
The actor is updated to maximise the scalarised critic output, enabling preference-conditioned policy learning under CTDE.}
    \label{algorithm_flow}
\end{figure}
\clearpage

\noindent where the noise term $\epsilon$ varies depending on the algorithm:

\begin{itemize}
    \item \textbf{MOMA-TD3:} Gaussian noise is clipped before being added to the target action:
    \begin{equation}
    \epsilon \sim \mathrm{clip}\bigl(\mathcal{N}(0, \sigma^2), -c, c\bigr).
    \end{equation}

    \item \textbf{MOMA-DDPG:} No noise is applied, as in the original DDPG formulation:
    \begin{equation}
    \epsilon = 0.
    \end{equation}
\end{itemize}

For MOMA-TD3, we compute the scalarised value from both critic targets.
\begin{equation}
Q^{(j)}_{\boldsymbol{\omega}}(o', a') = \boldsymbol{\omega}^{\!\top} \textbf{Q}_{\bar\theta_j}(o', a', \boldsymbol{\omega}), \qquad j \in \{1,2\},
\end{equation}

\noindent and select the critic estimate with the lower scalarised value:
\begin{equation}
j^\star = \arg\min_{j \in \{1,2\}} Q^{(j)}_{\boldsymbol{\omega}}(o', a'),
\label{bootstrap target}
\end{equation}
defining the bootstrap target vector as:
\begin{equation}\label{eq:critic_target}
\mathbf{y} = r + \gamma  \textbf{Q}_{\bar\theta_{j^\star}}(o', a',\boldsymbol{\omega}).
\end{equation}
For MOMA-DDPG, which uses only a single critic head ($j=j^\star=1$), the same form applies with:
\begin{equation}
\mathbf{y} = r + \gamma  \textbf{Q}_{\bar\theta_1}(o', a',\boldsymbol{\omega}) .
\end{equation}

\vspace{1ex}
\noindent Each critic head is trained using mean squared error:
\begin{equation}
\ell_j = \bigl\|\textbf{Q}_{\theta_j}(o, a, \boldsymbol{\omega}) - \mathbf{y}\bigr\|_2^2,
\end{equation}
and the overall critic loss becomes:
\begin{equation}
\label{eq:critic_loss}
\mathcal{L}_{\text{critic}} =
\begin{cases}
\ell_1 + \ell_2 & \text{(MOMA-TD3)} \\
\ell_1 & \text{(MOMA-DDPG)}.
\end{cases}
\end{equation}
This scalarised “minimum-of-two” target extends TD3’s overestimation bias reduction to the multi-objective setting.  
\Secref{sec:utility_overestimation} empirically validates that our dual-head architecture mitigates utility inflation, which is otherwise observed in single-critic setups.

\subsection{Multi-Headed Decentralised Actors}

Policy execution is decentralised, but computational cost is reduced via a shared actor trunk with agent-specific output heads. Each head maps its corresponding agent's local observation $o_i$ and utility preference $\boldsymbol{\omega}$ to an action $a_i = \pi_{i,\phi}(o_i, \boldsymbol{\omega})
$, using a shared trunk for feature extraction and backpropagation efficiency.

Actor training is guided by gradients from the centralised critic. For each agent $i$, we define a scalarised policy objective using the first critic:
\begin{equation}
J_i(\phi) =
\mathbb{E}_{(o, a_{-i}, \boldsymbol{\omega}) \sim \mathcal{D}} \left[
    \boldsymbol{\omega}^\top \textbf{Q}_{\theta_1} \big(o,( \pi_{i,\phi}(o_i,\boldsymbol{\omega}),a_{-i} ), \boldsymbol{\omega}\big)
\right],
\end{equation}
where $a_{-i}$ denotes the actions of all other agents stored in the replay buffer. This follows the structure of MADDPG's actor update, extended with utility scalarisation.

Using the deterministic policy gradient theorem, we compute the gradient:

\begin{equation}
    \nabla_{\phi} J_i =
\mathbb{E}_{(o, \boldsymbol{\omega}) \sim \mathcal{D}} \left[
    \nabla_{\phi} \pi_{i,\phi}(o_i,\boldsymbol{\omega})
    \nabla_{a_i} \left( \boldsymbol{\omega}^\top \textbf{Q}_{\theta_1}(o,a, \boldsymbol{\omega}) \right)
    \bigg|{a_i = \pi_{i,\phi}(o_i,\boldsymbol{\omega})}
\right].
\end{equation}
In practice, we minimise the negative scalarised utility as the actor loss:
\begin{equation}\label{eq:actor_loss}
\mathcal{L}_{\text{actor}}^i = -J_i(\phi),
\end{equation}
and optimise using Adam.

Concretely, given a minibatch $B=\{(o^k,a^k,\boldsymbol{\omega}^k)\}_{k=1}^N$ sampled from the replay buffer, we proceed as follows. For each sample $k$ and each agent $i$, form the joint action with agent $i$’s component replaced by its current policy output:
\[
\tilde a^k=\big(a_{-i}^k,\ \pi_{i,\phi}(o_i^{\,k},\boldsymbol{\omega}^k)\big),
\qquad
\text{where } a_{-i}^k=(a_1^k,\ldots,a_{i-1}^k,\,a_{i+1}^k,\ldots,a_n^k).
\]
We then evaluate the scalarised critic value $\boldsymbol{\omega}^{k\top} \textbf{Q}_{\theta_1}(o^k,\tilde a^k,\boldsymbol{\omega}^k)$ and accumulate its negative into the per-agent actor loss $\mathcal{L}_{\text{actor}}^i$. Losses are averaged over the batch, and all agent heads are updated via backpropagation through the shared trunk, allowing gradients from all agents to shape a common feature extractor while preserving decentralised execution at test time.


\begin{algorithm}
\caption{Training Procedure for MOMA-TD3
}
\label{pseudo-code: moma-td3}
\begin{algorithmic}[1]
\State Initialise environment $\mathcal{E}$ with $n$ agents
\State Initialise multi-headed actor network $\pi_\phi$
\State Initialise centralised critics $Q_{\theta_1},Q_{\theta_2}$
\State Initialise target networks:  $\pi_{\bar\phi}$, $Q_{\bar\theta_1}$ and $Q_{\bar\theta_2}$ 
\State Initialise replay buffers $\mathcal{D}_i$ for each agent $i$
\State Set hyperparameters: $\gamma$, $\tau$, $\sigma$, $\text{policy\_freq}$, $\text{batch\_size}$
\For{episode $=1$ to $M$}
    \State Sample preference vector $\omega_0 \sim \mathcal{U}(0,1)$
    \State Reset environment, receive initial observations $o=(o_1,o_2,...,o_n)$
    \While{episode not terminated}
        \For{each agent $i$}
            \State Pad $o_i$ to uniform dimension
            \State Select action $a_i = \pi_{i,\phi}(o_i, \boldsymbol{\omega}) + \epsilon_i$, $\epsilon_i \sim \mathcal{N}(0, \sigma^2)$
        \EndFor
        \State Execute joint action $a = (a_1, \dots, a_n)$ in $\mathcal{E}$
        \State Observe reward vector $\mathbf{r}$, next observation $o'$, and termination $d$
        \State Store $(o, a, \mathbf{r}, o', d, \boldsymbol{\omega})$ in buffers $\mathcal{D}_i$
        \State $o \leftarrow o'$
        \State Increment global timestep $t$
        \If{$t \mod \text{update\_freq} = 0$}
            \State Sample minibatch $\mathcal{B}$ from replay buffers
            \State Compute the joint next action using Equation \eqref{eq:actor_target}
            \State Compute $\mathbf{Q'_1}(o',a',\boldsymbol{\omega})$ and  $\mathbf{Q'_2}(o',a',\boldsymbol{\omega})$
            \State Compute critic target using Equation~\eqref{eq:critic_target}
            \State Update critic parameters $\theta_1,\theta_2$ by minimising Equation ~\eqref{eq:critic_loss}
            \If{$t \mod \text{policy\_freq} = 0$}
                \State Update multiheaded actor using Equation~\eqref{eq:actor_loss}: 
                \State Update target networks:
                \State \hspace{1em} $\bar\phi \leftarrow \tau \phi + (1 - \tau) \bar\phi$
                \State \hspace{1em} $\bar\theta_1 \leftarrow \tau \theta_1 + (1 - \tau) \bar\theta_1$
                \State \hspace{1em} $\bar\theta_2 \leftarrow \tau \theta_2 + (1 - \tau) \bar\theta_2$
            \EndIf
        \EndIf
    \EndWhile
\EndFor
\end{algorithmic}
\end{algorithm}


\subsection{Preference Sampling and Conditioning}

For each episode, a preference vector $\boldsymbol{\omega}$ is sampled uniformly from the unit simplex and held fixed for the episode. In our two-objective setting this reduces to $\omega_0 \sim \mathcal{U}(0,1)$ with $\omega_1 := 1-\omega_0$. We use a linear scalarisation function to collapse vector rewards into a single scalar. The sampled $\boldsymbol{\omega}$ conditions all actors and critics throughout the episode. This per-episode sampling encourages the policy to generalise across the preference space $\Omega$, yielding a single conditional policy that approximates a coverage set; we evaluate it using expected utility (under the same linear utility) and hypervolume computed over the induced Pareto front.

\subsection{Replay Buffer and Observation Padding}
\label{padding}
Each agent maintains a local replay buffer that stores tuples of the form $(o_i, a_i, \mathbf{r}_i, o'_i, d_i, \boldsymbol{\omega})$. For training, synchronised batches are sampled by timestep. To support shared networks despite heterogeneous observation spaces, each agent’s observation vector is zero-padded to a fixed maximum length.

\subsection{Training Protocol}

MOMA-TD3/DDPG is trained continuously over many episodes. A preference vector $\boldsymbol{\omega}$ is sampled at the start of each episode and held fixed. Agents interact with the environment using $\pi_i(o_i,\boldsymbol{\omega})$, generating experience tuples stored in per-agent buffers.

Training occurs at each timestep. The critic(s) are updated with Equation~\eqref{eq:critic_loss}, and the actor is updated using Equation~\eqref{eq:actor_loss}.  Exploration noise is added to actions during data collection, and stability is ensured via target networks, soft target updates and, in the case of MOMA-TD3, delayed actor updates.

At evaluation, agents act independently, conditioned on a user-specified preference vector $\boldsymbol{\omega}$. No central coordination is required at execution time.

\Algref{pseudo-code: moma-td3} provides pseudocode for MOMA-TD3.  
MOMA-DDPG is recovered by using a single critic head, setting $\texttt{policy\_freq} = 1$, and removing target policy noise.
A visual overview is shown in \Figref{algorithm_flow}.

\section{Experiments}
\label{sec:experiments}
In this section we detail the evaluation setup for MOMA-AC, including benchmark construction, baselines, and the metrics used to assess performance.

\subsection{Multi-Objective Multi-Agent MuJoCo}
The MuJoCo physics simulator is a standard platform in reinforcement learning, valued for its suite of continuous control tasks involving a range of robotic morphologies \cite{fujimoto2018addressing, khadka2018evolution, khadka2019collaborative}. In recent years, two orthogonal extensions of MuJoCo have emerged. First, a multi-objective version replaces the scalar reward functions with vector-valued signals that separately report metrics such as forward velocity and energy consumption. Second, MaMuJoCo introduces a multi-agent decomposition of robot morphologies, assigning control of different limb groups to distinct agents in a cooperative setting \cite{peng2021facmac}.

In this work, we integrate these two frameworks to construct a suite of cooperative, multi-objective, multi-agent environments. This combined setup enables rigorous evaluation of algorithms under realistic conditions that require decentralised control, continuous action spaces, and trade-offs across conflicting objectives.

\begin{figure}[h]
    \centering
    \includegraphics[width=\linewidth]{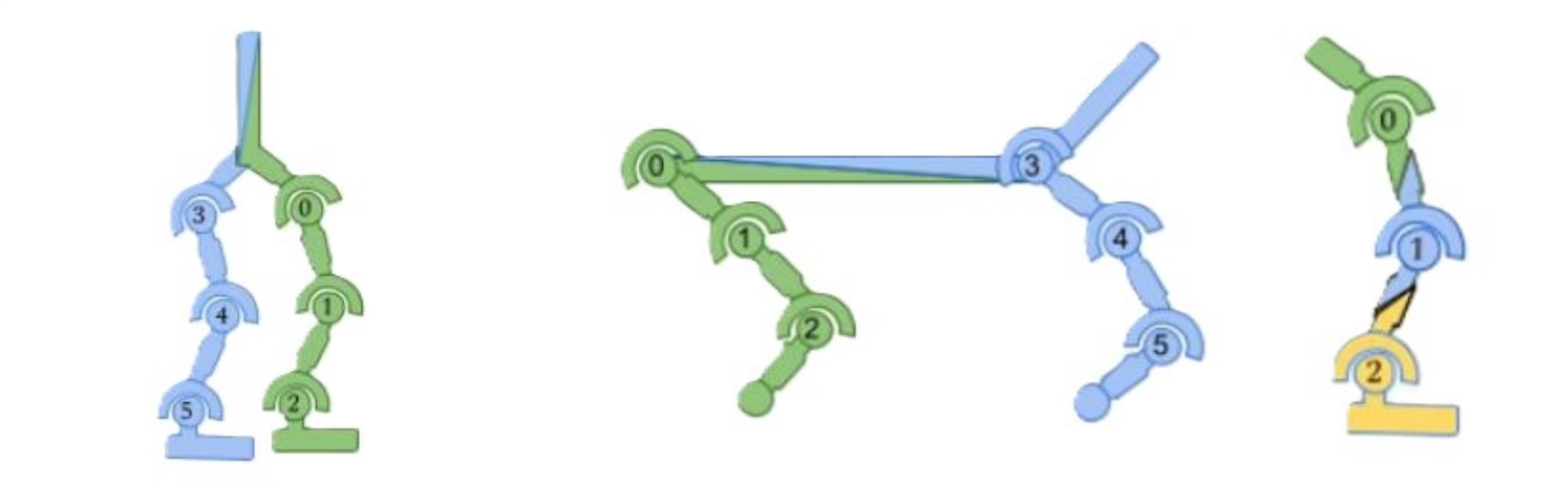}
    \caption{Walker(left), HalfCheetah (centre) and Hopper (right) morphologies in their decomposed form, adapted from the MaMuJoCo environment. Images sourced from the Farama Foundation documentation under the MIT License.}
    \label{fig:agents}
\end{figure}

We consider three base tasks—Hopper, HalfCheetah, and Walker, each with multiple agent decompositions. Hopper is run in single-agent and three-agent decompositions; HalfCheetah in single-, two-, and six-agent decompositions; and Walker in single- and two-agent decompositions\footnote{These are the only decompositions supported by the underlying MaMuJoCo package for these environments.}. This spectrum supports both baseline performance assessment but also scalability analyses as agent count increases. 

Each environment comprises a fixed set of actuated joints.  
A decomposition assigns these joints to mutually exclusive agents; if an agent controls more than one joint, as in the two-agent HalfCheetah, the joint-connectivity graph induced by the assignment is required to be connected.  
An agent’s action is a vector of torques bounded in \([-1,1]\), with length equal to the number of joints it controls.

Observations are local.  
An agent receives the positions and velocities of its own joints and of any joint, controlled by another agent, that lies within one kinematic link of its own set.  
This rule can yield unequal state dimensions: in the three-agent Hopper, \Figref{fig:agents}, the middle agent observes two neighbours, whereas the outer agents observe only one.  
To guarantee identical input sizes we zero-pad shorter observation vectors, as described in \Secref{padding}.

After acting, all agents receive the same two-component reward vector.  
For Hopper the components are forward velocity and vertical displacement; together they promote forward motion while encouraging occasional jumps.  
For HalfCheetah and Walker, the components are forward velocity and a control-cost penalty, balancing speed against actuation effort.

\subsection{Baseline}
We instantiate MOMA-AC with two widely used continuous-control algorithms, yielding MOMA-TD3 and MOMA-DDPG. The two variants share the same weight-conditioned multi-headed actors and centralised CTDE training; MOMA-TD3 retains TD3’s dual-critic, delayed policy updates, and target policy noise, while MOMA-DDPG removes these components (single critic, policy-freq=1, no target policy noise). 

We compare against two baselines:
\begin{itemize}
    \item \textbf{IND MO-TD3}: a decentralised inner-loop baseline where each agent trains its own weight-conditioned MO-TD3 policy without centralised critics or parameter sharing. (In the single-agent setting this is equivalent to MOMA-TD3.) 
    \item \textbf{Outer-Loop MA-TD3}: a scalarisation-first baseline that trains separate MA-TD3 policies over a grid of fixed preference weights, approximating a coverage set by solving multiple scalar tasks\footnote{Abbreviated in plots and tables to "OL MA-TD3".}.
\end{itemize}

\subsection{Key Metrics for Evaluation}
\label{metrics}

To evaluate our algorithms in the cooperative multi-agent setting, we adopt a suite of standard multi-objective performance metrics originally defined for single-agent contexts. In fully cooperative tasks, all agents share the same reward vector; thus, the Pareto front computed from any single agent's returns reflects the performance of the team as a whole. This enables us to leverage single-agent multi-objective metrics in a principled and interpretable manner.

\textbf{Expected Utility Metric (EUM)}:  
EUM quantifies a policy’s ability to generalise across diverse preferences by averaging scalarised utilities over a set of sampled weight vectors. Formally:

\[
\text{EUM} = \frac{1}{n} \sum_{i=1}^{n} U(\textbf{V}^{\pi(\cdot,\boldsymbol{\omega_i})},\boldsymbol{\omega_i}),
\]
where \( n \) is the number of sampled preference weight vectors \( \boldsymbol{\omega}_i \) and \( U(\textbf{V}^{\pi(\cdot,\boldsymbol{\omega_i})},\boldsymbol{\omega_i})\) denotes the utility under weight vector \( \boldsymbol{\omega}_i \). 

Higher EUM indicates greater preference-robustness. 
We set \( n = 50\) in all experiments.

\textbf{Hypervolume (HV)}:  
The hypervolume metric measures the volume in objective space that is weakly dominated by the set of predicted value vectors generated from our coverage set: \( \text{Cov}(\pi) \), and bounded by a predefined reference point. Formally:

\[
\text{HV}(\pi) = \lambda \left( \bigcup_{v \in \text{Cov}(\pi)} [v_1, ref_1] \times [v_2, ref_2] \right),
\]
where \( \lambda \) denotes the Lebesgue measure, and \( ref = (ref_1, ref_2) \) is the reference point. In our experiments, we use \( ref = (-1000, -1000) \).

A higher HV indicates that \( \text{Cov}(\pi) \) covers a larger and more optimal region of the objective space.

\textbf{Cardinality}:  
Cardinality measures the number of distinct Pareto-optimal policies identified by the algorithm, providing insight into the breadth of the explored trade-offs. It is defined as:

\[
C = |\mathcal{C}|,
\]
where \( \mathcal{C} \) is the set of non-dominated value vectors in the predicted coverage set. 

In settings with conflicting objectives, a higher cardinality indicates better exploration and representation of the Pareto front.

\textbf{Sparsity}:  
Sparsity reflects how uniformly the discovered solutions are distributed along the Pareto front. This is especially relevant when two methods yield similar hypervolumes. The sparsity metric is defined as:

\[
S = \frac{1}{n - 1} \sum_{d=1}^{m} \sum_{i=2}^{n} \left( p^{d}_{\text{sorted}}[i] - p^{d}_{\text{sorted}}[i-1] \right)^2,
\]
where \( n \) is the number of Pareto solutions, \( m \) the number of objectives, and \( p^{d}_{\text{sorted}} \) the list of values along objective \( d \), sorted in ascending order. Lower sparsity indicates better uniformity.

As is common in multi-objective reinforcement learning, no single metric provides a complete assessment of performance. For example, high cardinality paired with low EUM or HV may indicate a coverage set containing many diverse but low-quality solutions. Conversely, in settings with conflicting objectives, high EUM with low cardinality may reflect strong performance for specific preferences but limited coverage of the Pareto front. Similarly, two methods may exhibit similar HV values while differing significantly in sparsity, revealing discrepancies in the uniformity of their solution distributions. Taken together, these metrics offer a multifaceted evaluation of both the quality and diversity of learned policies across the preference space.

\section{Results}
\label{sec:results}
We evaluate the MOMA-AC framework, instantiated as MOMA-TD3 and MOMA-DDPG, on Hopper, HalfCheetah, and Walker across multiple agent decompositions. Each configuration is run for $500{,}000$ timesteps with $10$ random seeds. One timestep corresponds to the execution of a joint action.

Performance is evaluated every 100 episodes, with a final evaluation at $500{,}000$ timesteps. At each evaluation point we construct an approximate coverage set as follows: for each of 100 equally spaced preference weights on the unit simplex, the agents act deterministically for 5 episodes; we then average the vector returns across those 5 episodes. The set of undominated means forms the approximate Pareto front, and all metrics in \Secref{metrics} are computed on this front.

For visualisation purposes, Hopper and Walker curves are smoothed with a rolling average of window size $15$.\footnote{Smoothing reduces jaggedness caused by (i) variable episode lengths (and therefore number of episodes and evaluation points) in Hopper/Walker due to agents terminating early and (ii) seed-to-seed variance. All tables and hypothesis tests use raw (unsmoothed) data.
Because HalfCheetah supplies a constant number of timesteps per 100-episode interval, its curves appear naturally smoother than Hopper/Walker, where the effective timesteps per interval fluctuate.} To keep the presentation focused, 
Figures (\ref{fig:hc_results}), (\ref{fig:hp_results}) and (\ref{fig:wk_results}) display EUM and HV; Cardinality and Sparsity are summarised in Tables (\ref{tab:halfcheetah_results}), (\ref{tab:hopper_results}) and (\ref{tab:walker_results}).

All experiments were run on an Intel Core i9\textendash12900K CPU with 32\,GB of RAM and no GPU acceleration.

\subsection{Performance}
\begin{figure*}[h]
    \centering
    \includegraphics[width=1\linewidth,  trim=1cm 0cm 1cm 0cm,clip]{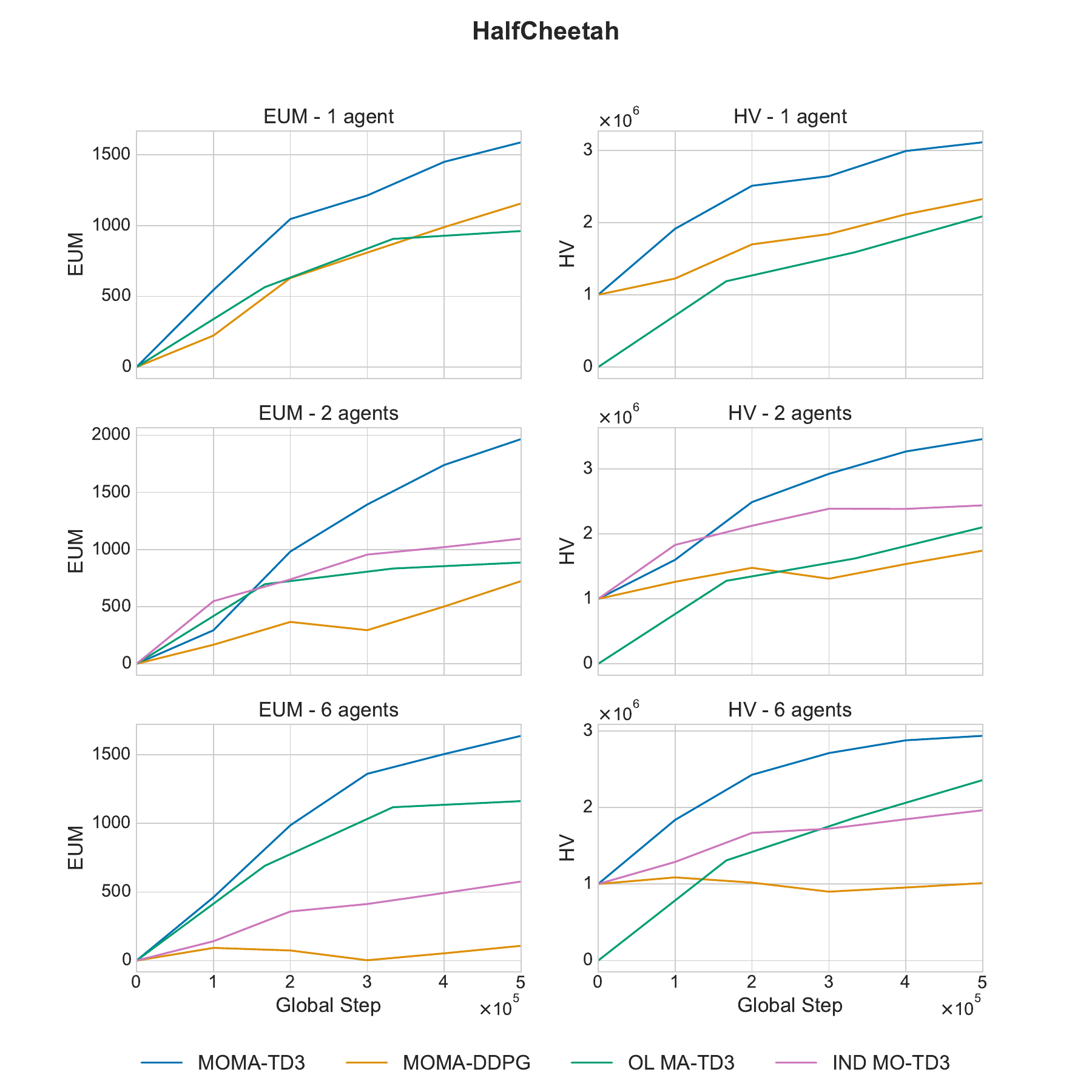}
    \caption{HalfCheetah (1, 2, 6 agents): EUM and HV learning curves comparing MOMA-TD3, MOMA-DDPG, IND, and Outer-Loop. MOMA-TD3 leads throughout, with gaps widening at higher agent counts.}

    \label{fig:hc_results}
\end{figure*}

\begin{figure*}[h]
    \centering
    \includegraphics[width=1\linewidth]{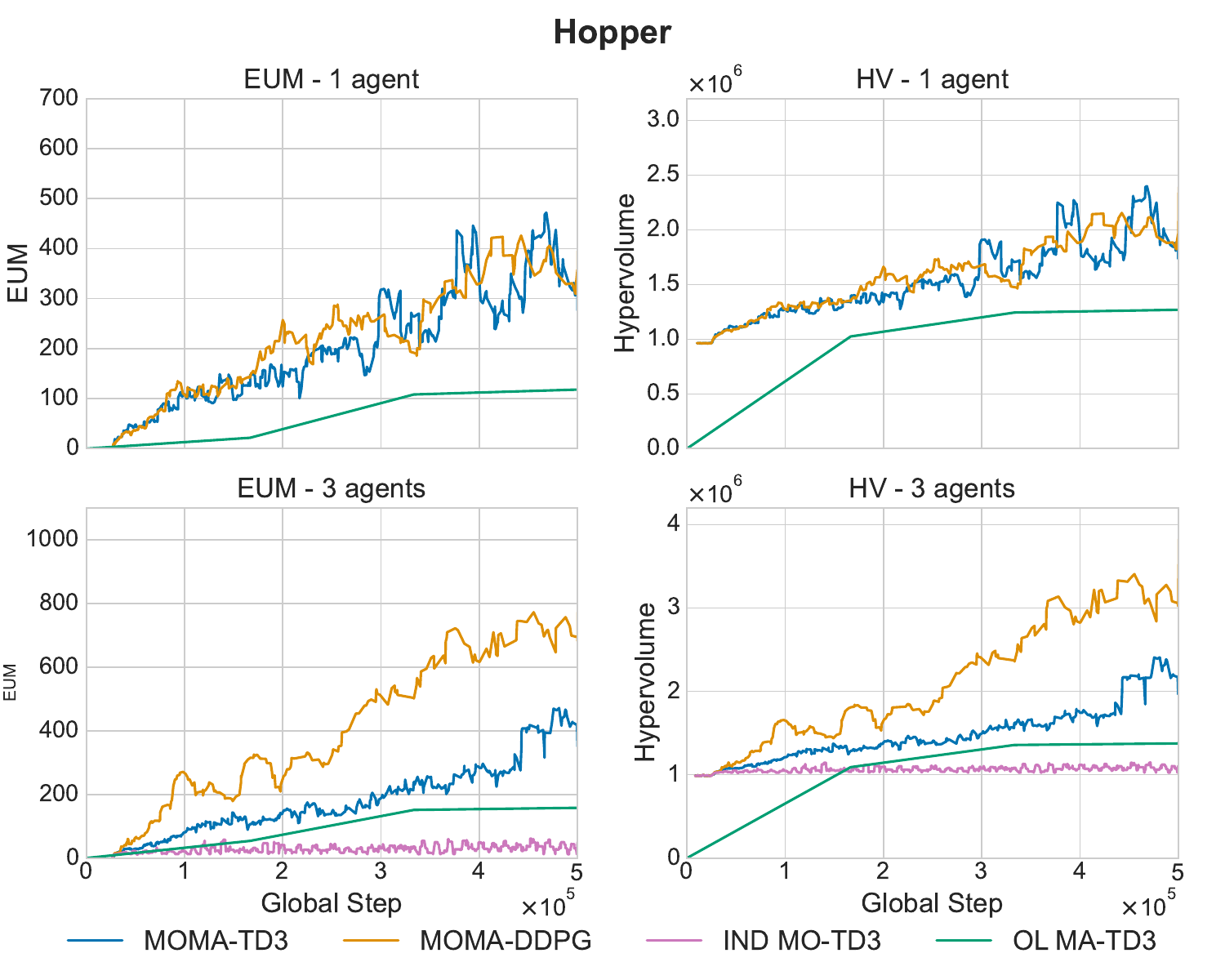}
    \caption{Hopper (1, 3 agents): EUM and HV learning curves. Both MOMA-AC variants beat baselines; MOMA-TD3 leads while MOMA-DDPG remains competitive.}

    \label{fig:hp_results}
\end{figure*}

\begin{figure*}[h]
    \centering
    \includegraphics[width=1\linewidth]{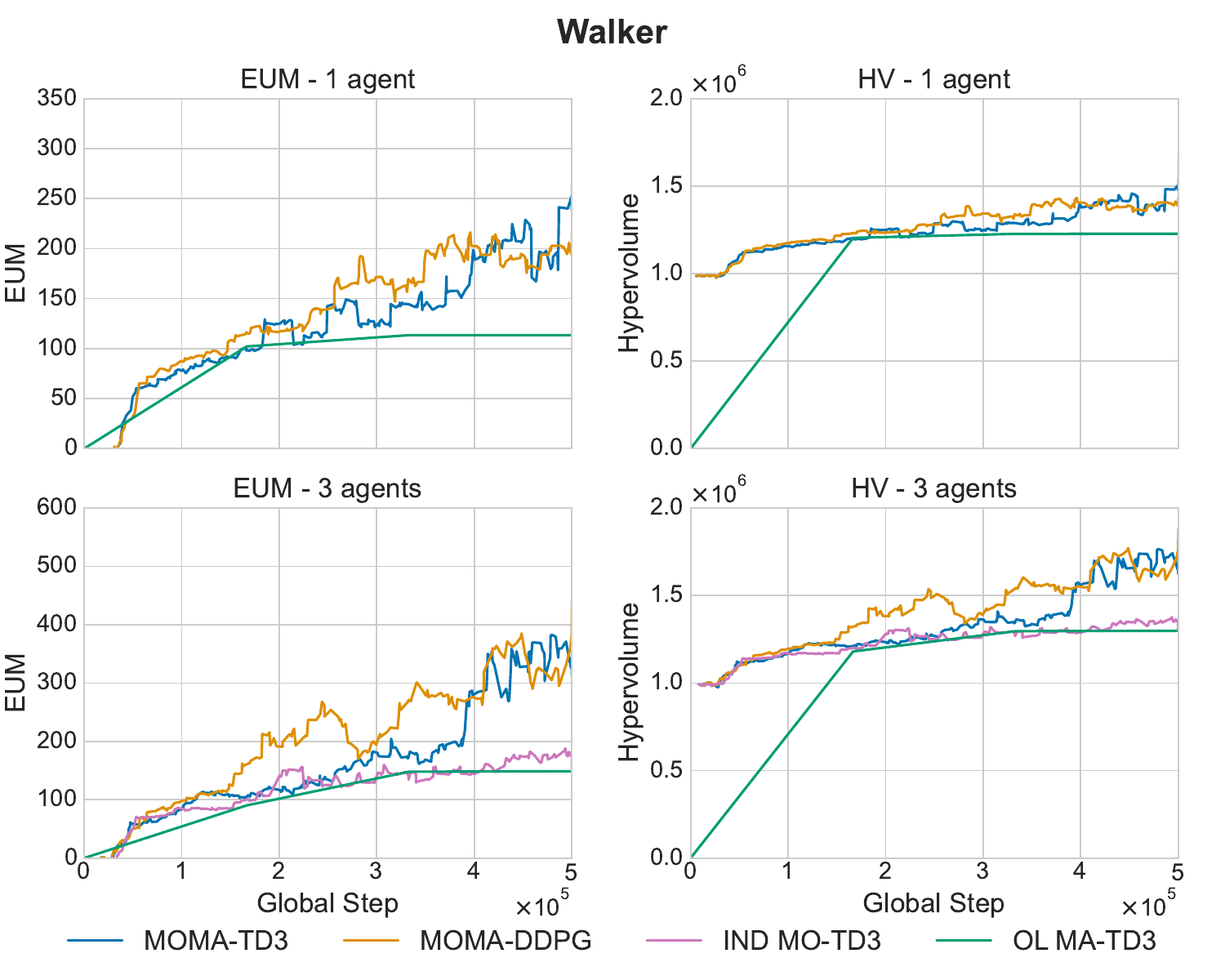}
    \caption{Walker (1, 2 agents): EUM and HV learning curves. Both MOMA-AC variants exceed baselines; MOMA-TD3 and MOMA-DDPG are comparable at these agent counts.}

    \label{fig:wk_results}
\end{figure*}

Across all environments and their multi-agent decompositions, the MOMA-AC framework delivers clear gains over both independent learners and the outer-loop baseline. In Figures (\ref{fig:hc_results}), (\ref{fig:hp_results}), and (\ref{fig:wk_results}), the EUM and HV curves for MOMA-TD3 separate early and remain higher than both throughout training, indicating consistently better trade-offs across preferences.

MOMA-DDPG performs competitively in Hopper and Walker, with no detectable difference from MOMA-TD3 at the $\alpha=0.05$ level (Welch’s unequal-variance two-sample t-test). In HalfCheetah, however, MOMA-TD3 pulls ahead as the number of agents increases: t-tests on EUM move from non-significant in the single-agent case ($p=0.18$) to strongly significant at two and six agents ($p=5.66\times 10^{-4}$ and $p=4.27\times 10^{-6}$, respectively). For HV, the same test finds MOMA-TD3 significantly better in all three HalfCheetah decompositions.

Counting significant wins across configurations (EUM), MOMA-TD3 improves over Outer-Loop in $6/7$ cases and over Independent learners in all $4/4$ multi-agent cases, while MOMA-DDPG achieves $4/7$ and $2/4$, respectively.\footnote{IND MO-TD3 has three fewer configurations than Outer-Loop as it is algorithmically indifferent to MOMA-TD3 in the single agent scenarios.}

\subsection{Coverage quality and diversity}
Cardinality and sparsity are informative only when read alongside quality. As Tables (\ref{tab:halfcheetah_results}), (\ref{tab:hopper_results}), and (\ref{tab:walker_results}) show, the Independent MO-TD3 baseline often matches (and occasionally exceeds) MOMA-TD3 in cardinality,and can exhibit lower sparsity in multi-agent HalfCheetah.  However, these extra points largely populate regions of the objective space that are dominated in utility compared to solutions found by MOMA-TD3/DDPG. In other words, IND MO-TD3 finds more solutions, but not more \emph{good} solutions.

By contrast, MOMA-TD3 couples high EUM/HV with comparable or higher cardinality, yielding coverage sets that are both broader and higher-valued across the preference simplex. The outer-loop baseline further underscores this point—its coverage sets are small by construction (one policy per fixed scalarisation) and concentrate mass at a few grid-preference optima, which is reflected in low cardinality.
Overall, the framework improves coverage that counts: more non-dominated solutions that deliver higher utility, rather than simply more points along a weaker front.

\subsection{Scalability}
A central requirement for any multi-agent method is stable performance as the number of agents increases. We assess scalability by comparing each multi-agent decomposition to its single-agent counterpart within the same environment. Again two-sided Welch t-tests are used (metrics defined in \Secref{metrics}).

MOMA-TD3 is robust to decomposition across all three tasks: there is no statistically significant change in either EUM or HV when moving from one to two or six agents in HalfCheetah, from one to three agents in Hopper, or from one to two agents in Walker. These results indicate that preference-conditioned CTDE with dual critics maintains Pareto quality as coordination complexity grows. 

MOMA-DDPG is more sensitive to agent count. In HalfCheetah, performance degrades with heavy decomposition: both EUM and HV are significantly lower at six agents than at one ($p{=}0.0010$ and $p{=}7.83{\times}10^{-4}$). In Hopper, EUM actually improves at three agents over the single agent variant (significant), while HV shows a non-significant trend. In Walker, both EUM and HV improve from one to two agents (both significant). 

\subsection{Utility overestimation}
\label{sec:utility_overestimation}

To disentangle framework effects from backbone algorithm effects, we measure utility overestimation for both MOMA-TD3 and MOMA-DDPG. 

For each trained actor–critic pair, we repeatedly sample a preference weight and a timestep. The agents are then allowed to act in the environment until that timestep, where the critic then predicts a vector return which we scalarise to a utility estimate. For MOMA-TD3 we take the minimum across its two critics. We then continue the episode with the current actor to obtain the realised (accumulated) utility from that timestep to termination. The utility error is
\[
e \;=\; \hat{U} - U,
\]

\noindent where negative values indicate conservative estimates and positive values indicate optimistic estimates.

\Tabref{tab:utility_overestimation} summarises the results. Across all three environments, MOMA-TD3 exhibits a conservative bias (negative mean error), which is statistically significant in HalfCheetah and Hopper. MOMA-DDPG shows an optimistic bias (positive mean error), significant in Hopper and Walker, and large but variable in HalfCheetah (non-significant with a high standard deviation). This pattern is consistent with the intended effect of dual critics and aligns with the main results: the TD3 instantiation remains stable as agent count increases, while the single-critic instantiation is more prone to optimistic utility estimates and degraded EUM/HV on harder decompositions. Because both variants share the same MOMA-AC machinery and protocol, these differences are attributable to the backbone algorithms rather than the framework itself.

\clearpage


\begin{landscape}

\begin{table}[h]
\centering
\caption{Mean (standard deviation) of evaluation metrics for HalfCheetah across different configurations. Metrics are averaged over 10 runs. MOMA-AC scalability is assessed across agent counts; MOMA-TD3/DDPG vs baseline comparisons are valid within matching decompositions.}

\label{tab:halfcheetah_results}
\begin{tabular}{lccccc}
\hline
\textbf{Config.} & \textbf{EUM} & \textbf{HV} & \textbf{Cardinality} & \textbf{Sparsity} \\
\hline
MOMA-TD3 1     & 1588.99 (682.15) & 3.11e+06 (7.57e+05) & 20.1 (8.27) & 2.58e+05 (3.12e+05) \\
MOMA-DDPG 1    & 1156.72 (710.67) & 2.33e+06 (8.54e+05) & 18.6 (9.25) & 9.36e+04 (1.25e+04) \\
Outer Loop 1  & 961.68 (344.11)  & 2.09e+06 (2.92e+05)  & 2.5 (0.53)  & 4.12e+06 (3.3e+06)\\
MOMA-TD3 2     & \textbf{1966.52 (260.62)} & 3.45e+06 (2.12e+05) & 20.5 (6.43) & 3.05e+05 (1.81e+05)  \\
IND 2      & 1095.29 (385.89) & 2.44e+06 (3.68e+05) & 20.6 (9.88) & 1.53e+05 (1.54e+05) \\
MOMA-DDPG 2   & 724.09 (778.32) & 1.74e+06 (9.61e+05) & 13.6 (7.29)  & 5.95e+04 (6.93e+04)  \\
Outer Loop 2 & 886.52 (283.46) & 2.11e+06 (2.9e+05)  & 2.8 (0.42) & 3.19e+06 (3.01e+06) \\
MOMA-TD3 6     & 1638.05 (586.89) & 2.94e+06 (6.15e+05) & 19.7 (7.86) & 4.73e+05 (5.65e+05) \\
IND 6      & 576.44 (237.45) & 1.96e+06 (2.65e+05) & 21.3 (8.12) & 5.88e+04 (7.77e+04)  \\
MOMA-DDPG 6   & 108.11 (341.32) & 1.01e+06 (4.59e+05) & 8.5 (4.72) & 1.63e+04 (1.77e+04) \\
Outer Loop 6 & 1162.14 (336.68) & 2.36e+06 (2.91e+05) & 2.3 (0.48) & 7.2e+06 (4.67e+06) \\

\hline
\end{tabular}
\end{table}

\begin{table}[h]
\centering
\caption{Mean (standard deviation) of evaluation metrics for Hopper across different configurations, averaged over 10 runs. MOMA-AC scalability is assessed across agent counts; MOMA-TD3/DDPG vs baseline comparisons are valid within matching decompositions.}
\label{tab:hopper_results}

\begin{tabular}{lccccc}
\hline
\textbf{Config.} & \textbf{EUM} & \textbf{HV} & \textbf{Cardinality} & \textbf{Sparsity} \\
\hline
MOMA-TD3 1  & 454.98 (332.17)     & 2.297e+06 (1.116e+06) & 10.50 (6.22)        & 5.244e+04 (9.332e+04) \\  
MOMA-DDPG 1 & 448.667(242.118677)        & 2.22e+6 (7.65e+05) &   6.5 (4.58)   & 3.52e+04 (4.9e+04)\\
Outer Loop 1 & 118.076 (86.21710879) & 1.27e+06 (1.85e+05) & 1.8 (0.63)  & 4.52e+04 (4.41e+04) \\
MOMA-TD3 3  & 447.33 (224.99)      & 2.193e+06 (6.239e+05) &  8.50 (8.86)        & 1.900e+04 (1.829e+04) \\
MOMA-DDPG 3 & \textbf{767.16 (380.95)} & 3.34e+06 (1.5e+06)  & 3.5 (2.12)   &  1.93e+05 (2.89e+05) \\
Outer Loop 3 & 157.955 (73.37) & 1.38e+06 (1.65e+05)  & 2.1 (0.57) & 7.67e+04 (1.11e+05) \\
IND 3   & 101.94 (118.34)      & 1.245e+06 (2.873e+05) &  9.00 (9.54)        & 2.884e+03 (4.712e+03) \\ 
\hline
\end{tabular}
\end{table}

\begin{table}[h]
\centering
\caption{Mean (standard deviation) of evaluation metrics for Walker across different configurations, averaged over 10 runs. MOMA-AC scalability is assessed across agent counts; MOMA-TD3/DDPG vs baseline comparisons are valid within matching decompositions.}
\label{tab:walker_results}
\begin{tabular}{lccccc}
\hline
\textbf{Config.} & \textbf{EUM} & \textbf{HV} & \textbf{Cardinality} & \textbf{Sparsity} \\
\hline
MOMA-TD3 1     & 301.699 (299.394) & 1.603e+06 (5.982e+05) & 20.6 (16.88) & 7.042e+02 (1.068e+03) \\  
MOMA-DDPG 1    & 197.973 (55.218)  & 1.396e+06 (1.104e+05) & 13.4 (3.72)  & 2.033e+03 (2.732e+03) \\  
Outer Loop 1   & 113.481 (26.183)  & 1.228e+06 (5.242e+04) & 2.8 (0.42)   & 2.480e+04 (1.644e+04) \\  
MOMA-TD3 2     & \textbf{446.495 (338.218)} & 1.867e+06 (6.916e+05) & 32.1 (12.61) & 1.265e+03 (2.366e+03) \\  
MOMA-DDPG 2    & 417.234 (247.892) & 1.835e+06 (4.956e+05) & 22.6 (9.72)  & 2.542e+03 (1.951e+03) \\  
Outer Loop 2   & 148.885 (146.658) & 1.298e+06 (2.932e+05) & 2.7 (0.67)   & 6.749e+04 (1.357e+05) \\  
IND 2          & 195.676 (82.846)  & 1.391e+06 (1.656e+05) & 13.8 (6.3)   & 7.028e+01 (9.878e+01) \\  
\hline
\end{tabular}
\end{table}
\end{landscape}

\begin{table}[h]
\centering
\caption{Utility estimation error statistics for dual-critic (MOMA-TD3) and single-critic (MOMA-DDPG) variants across all environments. One-sided t-tests are performed reflecting the directional hypothesis of conservative (MOMA-TD3) or optimistic (MOMA-DDPG) bias against the null hypothesis of zero mean error.}
\label{tab:utility_overestimation}
\begin{tabular}{lccc}
\hline
\textbf{Configuration} & \textbf{Mean} & \textbf{Std Dev} & \textbf{p-value} \\
\hline
MOMA-TD3 HalfCheetah  & $-37.61$  & $55.91$  & $< 0.0001$ \\
MOMA-DDPG HalfCheetah & $1.2\times10^4$  & $3.4\times10^4$  & $< 0.14$ \\
MOMA-TD3 Hopper       & $-3.72$   & $14.83$  & $< 0.0001$ \\
MOMA-DDPG Hopper      & $62.6$  & $87.74$  & $< 0.025$ \\
MOMA-TD3 Walker    & $-3.19$
& $16.99$  & $<0.28$ \\
MOMA-DDPG Walker & $23.39$ &
$39.32 $  & $<0.046$ \\
\hline
\end{tabular}
\end{table}

\subsection{Discussion}
The results support three conclusions. First, inner-loop, preference-conditioned CTDE reliably improves Pareto quality over both independent learners and outer-loop training, as evidenced by consistent separation in EUM/HV curves and significant final-score gains. Second, the framework scales: MOMA-TD3 preserves performance under finer decompositions across morphologies, demonstrating that centralised critics and shared, weight-conditioned actors provide the coordination signals needed for stable multi-agent learning. Third, conservative critic design matters: dual critics reduce optimistic utility bias and yield the most robust instantiation of the framework, especially as coordination complexity grows. Together, these findings establish MOMA-AC as a practical approach for learning diverse, high-quality trade-offs in continuous multi-objective multi-agent control.

\section{Future Work}
\label{sec:future_work}

MOMA-AC establishes a strong framework baseline with clear performance benefits over non-specialised approaches (independent learners and outer-loop training). Several research directions remain open.

\subsection{Generalised Multi-Agent Settings}

This work focuses on fully cooperative scenarios, in part due to limited simulators for continuous-state, continuous-action multi-objective multi-agent RL. Our integrated MOMA-MuJoCo testbed enforces full cooperation among agents. Two related avenues follow: (i) developing benchmarks that support cooperative, competitive, and mixed-motive interactions under vector-valued rewards; and (ii) extending the MOMA-AC framework to these broader settings—for example, by pairing centralised, weight-conditioned critics with decentralised actors that incorporate opponent modelling or communication, and by adapting evaluation to equilibrium notions appropriate for mixed incentives.

\subsection{Evolutionary Reinforcement Learning Extensions}

Evolutionary reinforcement learning (ERL)—combining policy-gradient methods with population-based search—has shown promise in both multi-agent and multi-objective RL. Instantiating MOMA-AC within an ERL paradigm is a natural next step: populations of preference-conditioned actors (and their centralised critics) could improve front coverage and training stability. A key open question is how to design evolutionary operators that respect both multi-objective dominance and multi-agent coordination, so that diversity translates into better Pareto trade-offs rather than merely more points.

\subsection{Many-Objective Reinforcement Learning and Improved Utility Sampling}
Our current setup samples preferences uniformly on the unit simplex and uses linear scalarisation. While effective with two objectives, this becomes inefficient as the number of objectives increases (many-objective RL, MaRL) or when the Pareto front is non-convex, where uniform draws can overemphasise extremes and leave gaps.

Future work should explore structured and adaptive samplers over $\Omega$—e.g., simplex-lattice or low-discrepancy designs for baseline coverage; coverage-aware or uncertainty-guided sampling to target under-represented or high-error regions; and curricula that progress from extreme weights to the interior to stabilise MaRL training. Beyond linear scalarisation, richer utilities (Chebyshev/achievement functions, reference-point or risk-aware objectives, or learned non-linear scalarisers) could better capture complex trade-offs; for MaRL evaluation,

However this is a multi-faceted research question due to the scarcity of benchmark environments that couple many objectives with multi-agent continuous control; developing such test-beds is a necessary companion effort for systematic progress in MaRL.

\section{Conclusion}
\label{sec:conclusion}

This work introduces MOMA-AC, to our knowledge the first dedicated inner-loop actor–critic framework for continuous multi-objective multi-agent reinforcement learning. The framework combines preference-conditioned centralised critics with decentralised multi-headed actors under CTDE, and we demonstrate two concrete instantiations, MOMA-TD3 and MOMA-DDPG.

Empirically, MOMA-AC consistently improves Pareto quality over independent learners and outer-loop baselines across Hopper, HalfCheetah, and Walker, as evidenced by sustained gains in Expected Utility and Hypervolume. The framework scales with agent count: the TD3 instantiation maintains performance under finer decompositions, while the DDPG instantiation is more sensitive in the hardest settings. Our utility-bias analysis further indicates that dual critics mitigate optimistic utility estimates, providing a mechanistic explanation for the robustness of the TD3 variant within the framework.

Together, these results position MOMA-AC as a practical foundation for preference-aware learning in continuous multi-agent systems, and chart a path toward more general, robust multi-objective architectures spanning mixed-motive interactions, richer preference models, and additional backbone algorithms.

\section{Acknowledgments}
This publication has emanated from research conducted with the financial
support of Taighde Eireann – Research Ireland under Grant No. 18/CRT/6223.

\bibliographystyle{unsrt}  
\bibliography{references}

\end{document}